\documentclass[sigconf,authorversion,nonacm]{acmart}

\usepackage{times}
\usepackage{helvet}
\usepackage{courier}
\usepackage{amsmath,amsfonts,amsthm}
\usepackage{float, graphicx, multirow}
\usepackage{makecell} 
\usepackage{hyperref}
\usepackage{lipsum}
\usepackage[toc,page]{appendix}
\usepackage{caption}
\usepackage{subcaption}

\AtBeginDocument{%
  \providecommand\BibTeX{{%
    \normalfont B\kern-0.5em{\scshape i\kern-0.25em b}\kern-0.8em\TeX}}}

\begin{document}

\title{Signed Directed Graph Contrastive Learning with Structure and Laplacian Augmentation}


\author{Taewook Ko}
\email{taewook.ko@snu.ac.kr}
\affiliation{%
  \institution{Seoul National University}
  \country{Republic of Korea}
}

\author{Yoonhyuk Choi}
\email{younhyuk95@snu.ac.kr}
\affiliation{%
  \institution{Seoul National University}
  \country{Republic of Korea}
}

\author{Chong-Kwon Kim}
\email{ckim@kentech.ac.kr}
\affiliation{%
  \institution{Korea Institute of Energy Technology}
  \country{Republic of Korea}
}

\renewcommand{\shortauthors}{Taewook Ko, et al.}

\begin{abstract}
Graph contrastive learning has become a powerful technique for several graph mining tasks. It learns discriminative representation from different perspectives of augmented graphs. Ubiquitous in our daily life, singed-directed graphs are the most complex and tricky to analyze among various graph types. That is why singed-directed graph contrastive learning has not been studied much yet, while there are many contrastive studies for unsigned and undirected. Thus, this paper proposes a novel signed-directed graph contrastive learning, SDGCL. It makes two different structurally perturbed graph views and gets node representations via magnetic Laplacian perturbation. We use a node-level contrastive loss to maximize the mutual information between the two graph views. The model is jointly learned with contrastive and supervised objectives. The graph encoder of SDGCL does not depend on social theories or predefined assumptions. Therefore it does not require finding triads or selecting neighbors to aggregate. It leverages only the edge signs and directions via magnetic Laplacian. To the best of our knowledge, it is the first to introduce magnetic Laplacian perturbation and signed spectral graph contrastive learning. The superiority of the proposed model is demonstrated through exhaustive experiments on four real-world datasets. SDGCL shows better performance than other state-of-the-art on four evaluation metrics.
\end{abstract}


%



\maketitle
\begin{sloppypar}

\section{Introduction}
Various types of online social graphs are developing as the use of the internet increases. Social graph research attracts attention from researchers with its usefulness for various downstream tasks such as prediction, classification, and recommendation\cite{fout2017protein,zhang2018end,fan2019graph,he2020lightgcn}. Signed-directed graphs, which express the relationship between users as like/dislike or trust/distrust, have the most information and are the most valuable among several graph types. However, it has difficulties to analyze them caused of severe noise and complexity.\\
Many network embedding and graph convolution studies for signed-directed graphs have been proposed based on the two well-known social theories, balance and status.\cite{heider1946attitudes,holland1971transitivity} Balance theory is the concept that "A friend of my friend is my friend, and an enemy of my friend is my enemy." Furthermore, status theory sets the relative user ranks with the relation of positive and negative edges. The ideas of most existing studies, including SiNE\cite{wang2017signed}, SGCN\cite{derr2018signed}, SNEA\cite{li2020learning}, and SDGNN\cite{huang2021sdgnn}, are derived from the two social theories. Recently, some studies that do not use them have been proposed. ROSE\cite{tong2020directed} proposed a user-role-based methodology by transforming a graph into an unsigned bipartite. SDGCN\cite{ko2022graph} proposed a spectral convolution via a signed magnetic Laplacian matrix. \\
Contrastive learning\cite{chen2020simple,he2020momentum} is a successful learning mechanism in computer vision research with fewer labels or a self-supervised environment. It learns discriminative representations through contrastive loss from the data itself. They maximize the mutual information of representations of augmented images. Recently, many graph studies utilizing contrastive learning have been suggested. Graph contrastive learning models generate graph views in their own ways, such as edge or node deletion\cite{you2020graph}, addition\cite{zeng2021contrastive}, random-walk\cite{qiu2020gcc}, or attribute masking\cite{zhu2021graph}. JOAO\cite{you2021graph} and AD-GCL\cite{suresh2021adversarial} learn optimal perturbing distribution according to the dataset and SimGRACE\cite{xia2022simgrace} adds gaussian noise on the trained parameters for encoder perturbation. They define contrastive loss with the augmented representations of nodes or graphs. Graph contrastive learning also aims to maximize the mutual information of different graph views. However, not much research has been conducted on signed-directed graphs. \\
SGCL\cite{shu2021sgcl} is the first study to apply contrastive learning on signed-directed. They make two graph views by giving random perturbations on edge signs and directions. Then the views are divided into positive graph views and negative graph views. The positive graph view is constructed with only the positive edges of a graph view, and the same for the negative graph view. Then run graph convolution on the different graph views to get augmented node representations. Though SGCL is innovative, there are some limitations. First, it loses the context of the original graph if we separate the edges into different graphs by the edge types. The mutual influence between edge types cannot be trained. For example, valuable patterns like the balance theory cannot be reflected. Second, most real-world signed graphs have a nearly 90\% of positive edge ratio\cite{kumar2016edge}. Therefore, the negative views are unsuitable for graph convolutions due to the sparsity issue. In this study, we propose a Signed Directed Graph Contrastive Learning(SDGCL), which improves these limitations. \\
SDGCL utilizes two levels of perturbation. We apply random edge perturbation similar to SGCN. It changes edge signs and directions to generate augmented graph views. Then we introduce magnetic Laplacian perturbation. It changes a phase parameter $q$ of the magnetic Laplacian matrix. $q$ controls the phase angle between the real- and imaginary axis\cite{ko2022graph}. Magnetic Laplacian was studied in the field of quantum physics\cite{shubin1994discrete,olgiati2017remarks,colin2013magnetic}. Thanks to its Hermitian properties, it has been used in graph convolutions for directed and signed-directed\cite{zhang2021magnet,singh2022signed}. Signed magnetic Laplacian encodes graph sturucture, including edge signs and directions. It reflects the graph information and allows the model to learn the patterns of nodes and edges. We define a spectral convolution layer with the perturbed Laplacian by the idea of graph signal processing\cite{defferrard2016convolutional,hammond2011wavelets,kipf2016semi}. SDGCL gets the two augmented node representations with the graph encoder and defines node-level contrast loss\cite{zhu2021graph,shu2021sgcl} for contrastive learning. \\
The performance of the proposed model was evaluated with the link sign prediction task, which is widely selected to evaluate the signed-directed graph embedding. It was tested with four real-world graphs and showed excellent performance compared to various baselines for signed-directed graph embedding and graph contrast learning.

The followings are the contributions of this paper
\begin{itemize}
\item This paper proposes a novel signed-directed graph contrastive learning model, SDGCL.
\item It augments node representations via two stages, graph structure perturbation and magnetic Laplacian perturbation.
\item To the best of our knowledge, it is the first to introduce magnetic Laplacian perturbation and is the first spectral signed graph contrastive model.
\item The proposed model has the best link sign prediction performance in real-world graph experiments.
\end{itemize}

\section{Related Works}
\subsection{Signed-Directed Graph Embedding}
Most signed-directed graph studies utilize sociological stories.\cite{heider1946attitudes, holland1971transitivity} SiNE\cite{wang2017signed} is an undirected model that uses balance theory, defining an objective function: friend nodes increase their similarity. BESIDE\cite{chen2018bridge} focused on the bridge edges to overcome the limitation of triads and utilized both social theories. SGCN\cite{derr2018signed} defined a novel balanced and unbalanced path for neighbor aggregation. SNEA\cite{li2020learning} and SiGAT\cite{huang2019signed} used attention mechanism. SDGNN defined four different weight matrices to distinguish the edge signs and directions. And they proposed triad loss based on the theories. Balance and status theories are developed with the patterns of the user triads. About 60-70\% of triads on real-world graphs satisfy the theories\cite{leskovec2010predicting,javari2017statistical}. Though the theories are crucial paradigms in signed-directed graph research, not all the triads satisfy them. Moreover, it is not a rule that all users should follow. Models that depend on the theories are not working well with users with little or no triads. On top of that, such models require much computational cost to find triads or paths. Recently, ROSE\cite{javari2020rose} proposed a methodology using user-role by transforming graphs into unsigned bipartite. SDGCN\cite{ko2022graph} proposed a spectral convolution model by proving the positive semidefinite of signed magnetic Laplacian.

\begin{figure}[t]
  \centering
  \includegraphics[width=\linewidth]{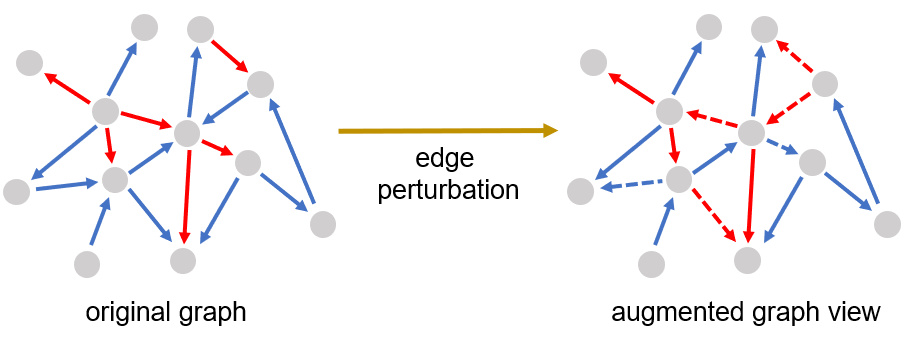}
  \caption{Example of a signed-directed graph and its structure perturbation. Dashed edges indicate perturbed edges.}
\end{figure}

\subsection{Magnetic Laplacian}
Magnetic Laplacian was first introduced in the field of quantum mechanics to analyze the charged particle under magnetic flux.\cite{shubin1994discrete,olgiati2017remarks,lieb1993fluxes,colin2013magnetic} But thanks to its Hermitian properties, the magnetic Laplacian has been utilized in directed graph studies.\cite{guo2017hermitian,liu2015hermitian}
Some application studies such as graph clustering\cite{cucuringu2020hermitian,cloninger2017note} node\cite{zhang2021magnet} and graph representation learning\cite{furutani2019graph} are delivered. MagNet\cite{zhang2021magnet} proved the positive semidefinite property of the magnetic Laplacian and proposed a spectral graph convolution. Recently, some studies extended the magnetic Laplacian to the signed graph. SDGCN\cite{ko2022graph} and SigMagNet\cite{fiorini2022sigmanet} define signed magnetic Laplacian. The signed magnetic Laplacian matrix uniquely encodes edge types of signed-directed graphs with real- and imaginary numbers. They proposed spectral graph convolution with the signed magnetic Laplacian.

\subsection{Graph Contrastive Learning}
SimCLR\cite{chen2020simple} and MoCo\cite{he2020momentum} proposed contrastive learning and achieved great success in image classification. Many follow-up studies were proposed thanks to the increasing learning efficiency via self-supervised. DGI\cite{velickovic2019deep}, GMI\cite{peng2020graph} adopted the contrastive learning to graph studies. They measure mutual information of input and node representations. InfoGraph\cite{sun2019infograph} proposed patch-level representation. GCC employed random-walk sampling to make positive and negative samples. Since then, several graph augmentations\cite{you2020graph,zeng2021contrastive,zhu2021graph} and graph encoders\cite{shu2021sgcl,xia2022simgrace} have been proposed for contrastive learning. There are contrastive studies for graph clustering\cite{pan2021multi,zhong2021graph}, node embedding\cite{zhu2021graph}, DDI prediction\cite{wang2021multi,li2022geomgcl}, and  recommendation\cite{lin2022improving}. However, signed-directed graph contrastive studies are not discussed much. To the best of our knowledge, SGCL\cite{shu2021sgcl} is the only one. SGCL makes graph views by randomly perturbing the edges. Then they construct positive- and negative-graph views by separating edges by the edge type. 
However, the edges of signed-directed graphs have meaningful relationships with each other. Therefore, the context of the data cannot be properly interpreted when we arbitrarily separate graph views by edge types. Moreover, they are inefficient in that they need to find positively or negatively linked nodes for every aggregation stage. To overcomes these limitations, this study proposes a novel graph contrastive learning model.

\section{Problem Formulation}
Let $\text{G} = (V, \mathcal{E}^+, \mathcal{E}^-)$ be a signed-directed graph where $V$ is a set of nodes like users in social graphs. $\mathcal{E}^+ \subseteq V \times V$ indicates positive edge matrix, and $\mathcal{E}^- \subseteq V \times V$ is of negative edges. Positive and negative edges represent user relationships, such as like/dislike or trust/distrust. For example $\mathcal{E}^+_{u,v}$ equals 1 if there is a positive edge from node $u$ to node $v$; otherwise, 0. Similarly, if there is a negative edge from node $v$ to $u$, $\mathcal{E}^-_{v,u}$ equals 1. Note that $\mathcal{E}^+ \cap \mathcal{E}^- = \varnothing$. Because we do not accept users having two different edges to the other user simultaneously, such as "John loves Jane and also hates her." A user may be connected to other users by one of the three relations(none, positive, negative). Thus, there are nine-edge types between a user pair. The goal of this paper is to map the nodes $u \in V$ into the low-dimensional embedding vectors $z_u \in \mathbb{R}^d$ with a given graph $\text{G}$ as:
\begin{displaymath}
    f(\text{G}) = Z,
\end{displaymath}
$f$ is a transformation function and $Z \in \mathbb{R}^{|V|\times d}$ is an embedding matrix. Each row of $Z$ represents the node embedding with dimension size $d$.

\section{Model Framework}
\subsection{Graph Augmentation}
\subsubsection{Structure Perturbation}
There are two methods for structure perturbation, edge sign perturbing and edge direction perturbing. In edge sign perturbing, we change the edge signs randomly from the given graph. For example, we sample $p\%$ of positive edges and change their signs to negative. And the same for the negative edges. Similarly, we sample $r\%$ of edges and change their directions to inverse. If an edge is bidirectional, we change it to directional arbitrary. With this random edge perturbation, we get structurally perturbed graph views, $\widetilde
{\text{G}} = (V, \tilde
{\mathcal{E}^+}, \tilde
{\mathcal{E}^+})$. Figure 1 shows an example of structure perturbation. As already known by balance and status theories, there is important information in the edge sign and directions. We may harm the vital triad patterns via random perturbation. However, we expect some amount of perturbation would be helpful to learn robust representations from noisy real-world data. Moreover, the perturbing can discover and exploit some relationships that might exist. The model can improve the generalization performance through random perturbations.

\subsubsection{Signed-Directed Magnetic Laplacian}
Before introducing Laplacian Perturbation, we define the signed-directed magnetic Laplacian. Graph Laplacian is useful to encode graph sturucture with degree and adjacency matrices, $\textbf{L}=\textbf{D}-\textbf{A}$. They are not only positive semidefinite but also have non-negative eigenvalues and associate orthonormal eigenvectors. With that properties, \cite{kipf2016semi} and \cite{hammond2011wavelets} proposed the idea of spectral graph convolution. However, graph Laplacian is asymmetric when a graph is directed or signed-directed. They have complex eigenvalues and do not satisfy the properties for spectral convolution. Thus, \cite{ko2022graph,singh2022signed,fiorini2022sigmanet} proposed a novel magnetic Laplacian matrix representing the structure of signed-directed graphs and satisfying positive semidefinite. First of all, we define a complex Hermitian adjacency matrix as follow,
\begin{equation*}
\mathbf{H}^q = \mathbf{A}_s \odot \mathbf{P}^{q}.
\end{equation*}
$\mathbf{A}_s := \frac{1}{2}(\mathbf{A}+ \mathbf{A^\intercal})\subseteq V \times V $ is a symmetrized adjacency matrix, and $\mathbf{P}^q \subseteq V \times V$ is a phase matrix with complex numbers. $\odot$ is an element-wise multiplication operation. The definition of the phase matrix is,
\begin{equation*}
\mathbf{P}^{q}(u,v) := \frac{\text{exp}(i\Theta_{uv}^{q})\text{A}_{uv}+ \text{exp}(i\overline{\Theta}^{q}_{uv})\text{A}_{vu}}
{|\text{exp}(i\Theta_{uv}^{q})\text{A}_{uv}+ \text{exp}(i\overline{\Theta}^{q}_{uv})\text{A}_{vu}|+ \epsilon}.
\end{equation*}
$\Theta_{uv}^{q} = q\mathcal{E}^+_{uv}+(\pi+q)\mathcal{E}^-_{uv}$ and 
$\overline{\Theta_{uv}}^{q} = -q\mathcal{E}^+_{vu}+(\pi-q)\mathcal{E}^-_{vu}$
Be careful with the subscripts orders. $q$ is a hyperparameter lies in [0,$\pi/2$]. The symmetrized adjacency matrix encodes the node connectivity, and the phase matrix encodes link directions and signs with different phase values. Figure 2(b) shows the edge encodings of the defined Hermitian adjacency matrix $\textbf{H}^q$. They uniquely encode the nine-edge types of signed-directed graphs. It tells node connections and edge types as well. We can see that $\textbf{H}^q$ is a complex numbered skew-symmetric form, a complex Hermitian matrix. Then, we define the signed-directed magnetic Laplacians with this Hermitian adjacency by,
\begin{equation*}
    \mathbf{L}^{q}_U := \mathbf{D}_s-
    \mathbf{H}^{q} = \mathbf{D}_s- \mathbf{A}_s\odot \mathbf{P}^q
\end{equation*}
\begin{equation*}
    \mathbf{L}^{q}_N := \mathbf{I}-(\mathbf{D}^{-\frac{1}{2}}_s \mathbf{A}_s \mathbf{D}^{-\frac{1}{2}}_s)\odot \mathbf{P}^q,
\end{equation*}
$\mathbf{D}_s$ is a symmetric degree matrix, similar to $\mathbf{A}_s$. $\mathbf{L}^{q}_U$ and $\mathbf{L}^{q}_N$ are unnormalized and normalized signed-directed magnetic Laplacians. It is well known that the skew-symmetric, complex Hermitian matrix is positive semidefinite\cite{ko2022graph,fiorini2022sigmanet,singh2022signed}. Thus, the Laplacians are positive semidefinite and diagonalizable by spectral decomposition. For example, the normalized Laplacian is diagonalized by,
\begin{equation*}
    \textbf{L}^{q}_N = \textbf{U}\Lambda\textbf{U}^{\dagger}.
\end{equation*}
Each column of $\mathbf{U}$ is eigenvector $\mathbf{u}_k$ and $\mathbf{U}^{\dagger}$ is a conjugate transpose of $\mathbf{U}$. $\Lambda$ is a diagonal matrix where the elements are  $k$-th eigenvalues $\Lambda_{k,k} = \mathbf{\lambda}_k$. The eigenvalues and eigenvectors contain the structural information of the signed-directed graph. We leverage this matrix to define spectral graph convolution.

\begin{figure}[t]
  \centering
  \includegraphics[width=\linewidth]{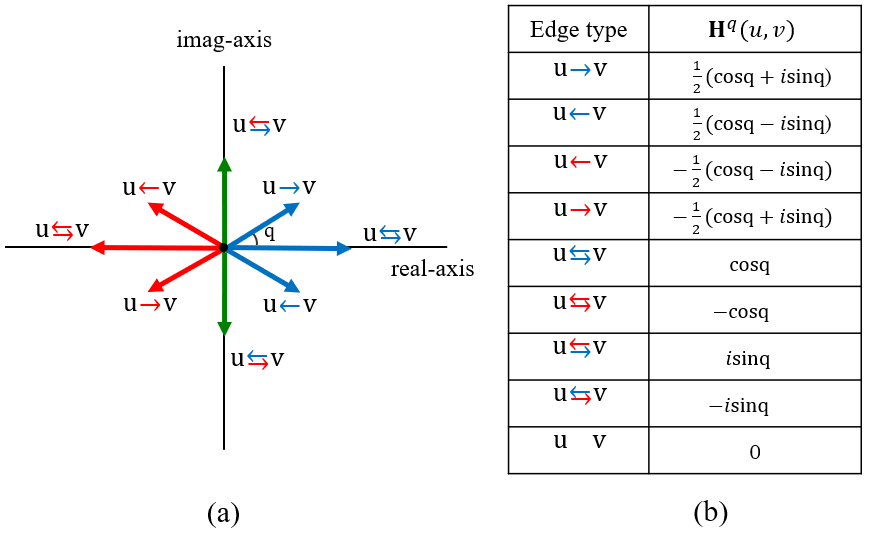}
  \caption{Edge encoding and meaning of $q$. It shows edge encoding values of a complex Hermitian adjacency matrix. $q$ controls the phase angle.}
\end{figure}

\begin{figure*}[t]
  \centering
  \includegraphics[width=\linewidth]{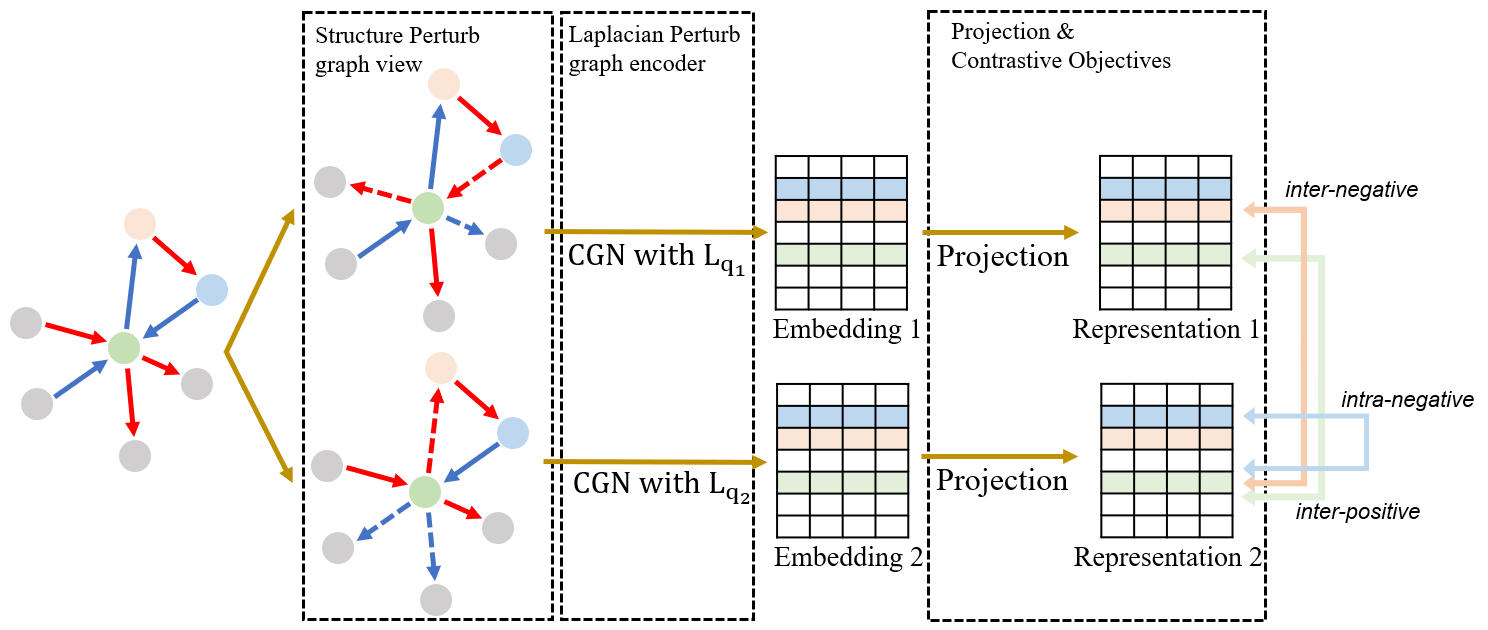}
  \caption{Model overview}
\end{figure*}

\subsubsection{Laplacian Perturbation}
Figure 2(a) describes the meaning of $q$ in magnetic Laplacian. $q$ controls the phase angle of encoded values between the real- and imaginary axes. If $q$ is small, the Laplacian puts less attention on directional information. It becomes an undirected model when the $q$ is 0. However, the high $q$ value also harms the encoding validity. $q$ influences the encoding effectiveness of the signed-directed magnetic Laplacian. In this subsection, we introduce Laplacian perturbation via $q$ variation. We randomly select $q$ value from 0 to 0.4$\pi$. The Laplacian matrix varies according to the $q$ variations, though the meaning of the graph structure that the Laplacian represents is still the same. This kind of Laplacian perturbation is a useful technique that makes graph augmentation without distortion of the original data\cite{tong2020digraph}.
\\In short, we make two structurally perturbed graph views $\widetilde{\textbf{G}}_1$ and $\widetilde{\textbf{G}}_2$, through random edge changes. Then we get perturbed signed-directed magnetic Laplacian $\widetilde{\textbf{L}}^{q_1}$ and $\widetilde{\textbf{L}}^{q_2}$ from the graph views with randomly chosen $q$ values. With these Laplacian matrices, we define graph encoders.

\subsection{Graph Encoder}
\subsubsection{Spectral Convolution via Magnetic Laplacian}
The signed-directed magnetic Laplacian $\mathbf{L}$, is diagonalizable with eigenvector matrix $\mathbf{U}$, and diagonal eigenvalues $\Lambda$. Several graph convolution studies \cite{defferrard2016convolutional,hammond2011wavelets} utilize the eigenvectors as discrete Fourier modes of graph signal processing. The graph signals are transformed into the Fourier domain through graph Fourier transform $\mathbf{x}:V \to \mathbb{C}$ by $\mathbf{\hat{\textbf{x}}}=\mathbf{U}^{\dagger}\mathbf{x}$. The inverse Fourier transform formula is defined as follow thanks to the unitarity of $\mathbf{U}$,
\begin{displaymath}
 \mathbf{x}=\mathbf{U}\mathbf{\hat{x}}= \sum_{k=1}^{N}\hat{\mathbf{x}}(k)\mathbf{u}_k. 
\end{displaymath}
The spectral convolution operation of graph signal processing is described as
\begin{equation*}
 \mathbf{g}_\theta \ast \mathbf{x} = \mathbf{U} \mathbf{g}_\theta \mathbf{U}^{\dagger}\mathbf{x},
\end{equation*}
where $\mathbf{g}_\theta=diag(\theta)$ is a trainable filter. For efficient calculation, \cite{hammond2011wavelets} proposed a truncated Chebyshev polynomial expansion of the filter by,
\begin{displaymath}
 \mathbf{g}_{\theta'}(\mathbf{\Lambda})\approx 
 \sum_{k=0}^{K}\theta'_k T_k(\mathbf{\tilde{\Lambda}}).
\end{displaymath} 
$T_0(x)=1, T_1(x)=x$, and $T_k=2x T_{k-1}(x)+T_{k-2}(x)$ for $k \geq 2$. $k$ is an expansion order. $\mathbf{\tilde{\Lambda}}=\frac{2}{\lambda_{max}}\mathbf{\Lambda}-\mathbf{I}$ is a normalized eigenvalue matrix, and $\lambda_{max}$ is the largest eigenvalue. $\theta'_k$ are Chebyshev coefficients. We have a simplified form of spectral graph convolution by,
\begin{equation*}
 \mathbf{g}_{\theta'} \ast \mathbf{x} = 
 \sum_{k=0}^{K}\theta'_k T_k(\mathbf{\tilde{L}})x,
 \end{equation*} 
where $\mathbf{\tilde{L}}=\frac{2}{\lambda_{max}}\mathbf{L}-\mathbf{I}$ analogous to $\mathbf{\tilde{\Lambda}}$. 

\subsubsection{Spectral Convolution Layer}
We define the spectral convolution layer with the approximated signed-directed spectral convolution operation. We set $k$ as 1, the maximum polynomial order, and $\lambda_{max}$ is assumed $2$ to make it practical. Similar to GCN\cite{kipf2016semi}, we set $\theta=\theta_0'=-\theta_1'$. Then we have    
\begin{equation*}
    \mathbf{g}_{\theta'} \ast \mathbf{x} \approx \theta(\mathbf{I}+(\mathbf{D}^{-\frac{1}{2}}_s \mathbf{A}_s \mathbf{D}^{-\frac{1}{2}}_s) \odot \mathbf{P}^q)\mathbf{x}.
\end{equation*}
The spectral convolution layer is defined as 
\begin{displaymath}
  \textbf{H} = 
  (\mathbf{\tilde{D}}^{-\frac{1}{2}}_s \mathbf{\tilde{A}}_s \mathbf{\tilde{D}}^{-\frac{1}{2}}_s\odot \mathbf{P}^q) \mathbf{X} \mathbf{W}.
\end{displaymath} 
$\textbf{H}\in \mathbb{R}^{N \times F}$ is the convoluted graph signals or representations. $\mathbf{X}\in \mathbb{R}^{N \times C}$ is the input signal. $C$ and $F$ are the numbers of input and output channels. $\textbf{W} \in \mathbb{R}^{C \times F}$ is a learnable matrix. It is a renormalization trick that $\mathbf{I}+(\mathbf{D}^{-\frac{1}{2}}_s \mathbf{A}_s \mathbf{D}^{-\frac{1}{2}}_s)\odot \mathbf{P}^q \to \mathbf{\tilde{D}}^{-\frac{1}{2}}_s \mathbf{\tilde{A}}_s \mathbf{\tilde{D}}^{-\frac{1}{2}}_s\odot \mathbf{P}^q$ where $\mathbf{\tilde{A}_s} = \mathbf{A}_s+\mathbf{I}$ and $\mathbf{\tilde{D}_s}(i,i)=\sum_j{\mathbf{\tilde{A}_s}(i,j)}$. It prevents gradient vanishing and exploding problems. 

\subsubsection{Signed-Directed Graph Encoder}
The graph encoder stacks $L$ layers of proposed spectral convolution layer. The $l$-th layer feature vector $\mathbf{x}^{(l)}$ is defined as,
\begin{displaymath}
  \mathbf{x}_j^{(l)}= \sigma ( \sum_{i=1}^{F_{l-1}} \mathbf{Y}_{ij}^{(l)}\mathbf{x}_i^{(l-1)}+\mathbf{b}_j^{(l)}).
\end{displaymath} 
We use a novel activation function for the complex elements. The activation function $\sigma$ is an complex version of ReLU, $\sigma(z)=z$, if $-\pi/2 \leq \text{arg}(z) \leq \pi/2$, otherwise, $\sigma(z)=0$. The feature matrix $\mathbf{X}^{(L)}$ has both real and imaginary values. Here we introduce unwinding operation. It converts real and imaginary features into the same domain.
\begin{displaymath}
    \mathbf{X}_{\text{unwind}}^{(L)} = 
[\text{real}(\mathbf{X}^{(L)}) ||
    \text{imag}(\mathbf{X}^{(L)}) \otimes (-i)].
\end{displaymath} 
We add a fully connected layer after unwinding to get the node representation. 
\begin{equation*}
    \textbf{Z} = \sigma(\mathbf{X}_{\text{unwind}}^{(L)}\mathbf{W}^{L+1} + \mathbf{B}^{(L+1)})
\end{equation*}
$\textbf{Z} \in \mathbb{R}^{N\times D}$ is the output node representation. The augmented two graph views are fed into the spectral graph encoder and makes the augmented node representations.

\subsection{Contrastive Objective}
There are two graph views after perturbations, and the graph encoder makes node representations of them $\textbf{Z}^1$ and $\textbf{Z}^2$. The goal of the contrastive objective is that the embeddings of the same nodes agree with each other. At the same time, they distinguish from the other nodes. We apply a MLP projection layer to improve the discriminative power.\cite{jacovi2021contrastive,chen2020big} We define the contrastive losses with the projected representation $\textbf{M}$.

\subsubsection{Inter-view Loss}
It is considered that the two identical nodes from different views are inter-positive pairs. On the other hand, the rest of the nodes are considered inter-negative pairs. For example, a node $u$ from graph view 1 and another node $u$ from graph view 2 are inter-positive pairs. And others nodes $v \in \text{V} $ and $v \neq u$ from graph view 2 are inter-negative pairs with a node $u$ from graph view 1. Even though the augmented representations from different graph views are different, they represent the same nodes. We want to maximize the agreement of the positive pair representations $\textbf{m}^{v_1}_u$ and $\textbf{m}^{v_2}_u$. While minimize the agreements of the negative pairs $\textbf{m}^{v_1}_u$ and $\textbf{m}^{v_2}_v$. The goal of the inter-view objective is to maximize the similarity of positive pairs and minimize the negative pairs. 
\begin{equation*}
  \mathcal{L}_{inter} = -\frac{1}{N}\sum^{N}_{i=i} \text{log}\frac{\text{exp}(sim(\textbf{m}^{v_1}_i, \textbf{m}^{v_2}_i)/\tau)}{\sum^{N}_{j=1,j \neq i}\text{exp}(sim(\textbf{m}^{v_1}_i,\textbf{m}^{v_2}_j)/\tau)}
\end{equation*}
$sim$ is a cosine similarity operation, and $\tau$ is a temperature parameter. 

\subsubsection{Intra-view Loss}
We define intra-view loss as similar to inter-view loss. The inter-view loss compares the projected node representations between the two different graph views. In contrast, inter-view loss calculates the discriminative loss within a graph view. Each node in a graph view is a unique user. All nodes have their representations, and we need to distinguish them from others. Therefore, we define inter-view loss as,  
\begin{equation*}
  \mathcal{L}_{intra} = -\frac{1}{N}\sum^{N}_{i=i} \text{log}\frac{1}{\sum^{N}_{j=1,j\neq i}\text{exp}(sim(\textbf{m}^{v}_i,\textbf{m}^{v}_j)/\tau)}.
\end{equation*}

\subsubsection{Contrastive Loss}
The whole contrastive loss is the sum of the inter- and intra-view loss functions. The goal of the contrastive loss is to let the model learn discriminative power from the projected node representations and maximize the positive agreements/mutual information. 
\begin{equation*}
  \mathcal{L}_{contrastive} = \mathcal{L}_{inter}+ \mathcal{L}_{intra}
\end{equation*}

\subsection{Prediction and Label Loss}
For model training, we not only use contrastive loss but also utilize label loss. The augmented two graph views are inputs of graph encoder and makes two node representations $\textbf{Z}^1$ and $\textbf{Z}^2$. The representations are concatenated and fed into output layer. The output layer makes the final node representation. 
\begin{equation*}
  \textbf{R} = \sigma([\textbf{Z}^1 \parallel \textbf{Z}^2] \textbf{W}^{out}+\textbf{B}^{out})
\end{equation*}
$\textbf{R} \subseteq V \times d $ and $\textbf{W}^{out} \subseteq 2d  \times d $ where $d$ is embedding dimension. We make prediction results with this output node representation. The prediction layer is defined as, 
\begin{equation*}
  \hat{y}_{u,v} = \sigma([\textbf{r}^1_u||\textbf{r}^2_v]{\textbf{W}}^{pred}+\textbf{B}^{pred})
\end{equation*}
Prediction value estimates the link sign when there is an edge from node $u$ to $v$. We define the label loss with the prediction error.
\begin{equation*}
  \mathcal{L}_{label} = -\sum_{{u,v}\in\mathcal{E}^{+}}^{|\mathcal{E}^+|} y_{u,v} \text{log}\hat{y}_{u,v} -\sum_{{u,v}\in\mathcal{E}^{-}}^{|\mathcal{E}^-|} (1-y_{u,v}) \text{log} (1-\hat{y}_{u,v}) 
 \end{equation*}
SDGCL is trained with the following objective function,
\begin{equation*}
  \mathcal{L} = \alpha \times \mathcal{L}_{contrastive}+ \mathcal{L}_{label}.
\end{equation*}
$\alpha$ is weight of contrastive loss. Note that, unlike other graph contrastive studies, it leverages supervised labels.

\section{Experiments}
\subsection{Datasets and Metrics}
We evaluate the model with four real-world signed directed graph datasets widely used in the link sign prediction task. Bitcoin-Alpha\footnote{
http://www.btc-alpha.com} and Bitcoin-OTC\footnote{http://www.bitcoin-otc.com}\cite{kumar2016edge} are extracted from Bitcoin trading platforms. Nodes are users, and edges are user relationships. Users can score the others on a scale of -10 to +10. Edges higher than 0 are treated as positive edges, otherwise negative edges. Epinions\footnote{http://www.epinions.com}\cite{guha2004propagation} is a who-trust-whom network crawled from a consumer review site. Users can notate trust or distrust to reviews of other users. Slashdot\footnote{http://www.slashdot.com}\cite{kunegis2009slashdot} is a social network of user community site. Especially they share new information. Users tag others as friends or foes, and we can construct positive and negative edges with this information. Moreover, we adopt four metrics, AUC, macro-F1, micro-F1, and binary-F1, for unbiased evaluation.

\begin{table}[h]
\centering
\renewcommand{\arraystretch}{1.3}
\resizebox{\columnwidth}{!}{%
\begin{tabular}{ccccc}
\Xhline{2.5\arrayrulewidth}
Dataset & \# nodes & \# pos links & \# neg links & positive ratio \\ \hline
Bitcoin-Alpha & 3,783 & 22,650 & 1,536 & 0.937 \\
Bitcoin-OTC & 5,881 & 32,029 & 3,563 & 0.900 \\
Epinions & 131,828 & 717,667 & 123,705 & 0.853 \\
Slashdot & 82,144 & 425,072 & 124,130 & 0.774 \\ \Xhline{2.5\arrayrulewidth}
\end{tabular}%
}
\caption{Dataset statistics.}
\label{tab:Dataset Statistics}
\end{table}

\begin{table*}[t]
\centering
\renewcommand{\arraystretch}{1.2}
\tiny
\resizebox{\textwidth}{!}{%
\begin{tabular}{cc|ccc|ccc|cccc}
\Xhline{2\arrayrulewidth}
&  & \multicolumn{3}{c|}{Signed Convolution} & \multicolumn{3}{c|}{Contrastive Learning} & \multicolumn{4}{c}{Signed Contrastive Learning}   \\ \hline
\multicolumn{1}{c|}{Dataset} & Metric & SGCN & SDGNN & SDGCN & GraphCL & GCA & SimGRACE & SGCL & SDGCL-\tiny{s} & SDGCL-\tiny{l} & SDGCL \\   \Xhline{2\arrayrulewidth}
\multicolumn{1}{c|}{\multirow{4}{*}{Bitcoin-Alpha}} & AUC & 0.782 & 0.835 & 0.858 & 0.814 & 0.838 & 0.823 & 0.849 & \textbf{0.896} & 0.883 & \underline{0.886} \\
\multicolumn{1}{c|}{} & Macro-F1  & 0.668 & 0.683 & 0.723 & 0.653 & 0.671 & 0.657 & 0.712 & 0.740 & \underline{0.744} & \textbf{0.754}  \\
\multicolumn{1}{c|}{} & Micro-F1  & 0.899 & 0.909 & 0.923 & 0.907 & 0.913 & 0.919  & 0.923 & \underline{0.947} & 0.942 & \textbf{0.949}\\
\multicolumn{1}{c|}{} & Binary-F1  & 0.941 & 0.947 & 0.958 & 0.950 & 0.953 & 0.957 &  0.959 & \textbf{0.973} & 0.969 & \underline{0.971}\\  \hline

\multicolumn{1}{c|}{\multirow{4}{*}{Bitcoin-OTC}} & AUC & 0.832 & 0.879 & 0.887 & 0.852 & 0.868 & 0.859 & 0.893 & \textbf{0.914} & 0.902 & \underline{0.910}\\   
\multicolumn{1}{c|}{} & Macro-F1  & 0.710 & 0.751 & 0.773 & 0.725 & 0.743 & 0.725 & 0.781 & \textbf{0.803} & \underline{0.796} & 0.802 \\
\multicolumn{1}{c|}{} & Micro-F1  & 0.886 & 0.902 & 0.911 & 0.904 & 0.907 & 0.906  & 0.920 & \underline{0.935} & 0.930 & \textbf{0.937}\\
\multicolumn{1}{c|}{} & Binary-F1  & 0.924 & 0.938 & 0.950 & 0.948 & 0.948 & 0.948 & 0.956 &  \underline{0.964} & 0.962 & \textbf{0.965}\\  \hline

\multicolumn{1}{c|}{\multirow{4}{*}{Epinions}} & AUC & 0.848 & 0.914 & 0.939 & 0.839 & 0.911 & 0.913 & 0.876 & 0.941 & \textbf{0.943} & \underline{0.942}\\   
\multicolumn{1}{c|}{} & Macro-F1  & 0.741 & 0.831 & 0.850 & 0.726 & 0.814 & 0.812 & 0.798 & 0.861 & \textbf{0.865} & \underline{0.863} \\
\multicolumn{1}{c|}{} & Micro-F1  & 0.893 & 0.912 & 0.925 & 0.887 & 0.913 & 0.915  & 0.909 & \underline{0.934} & \textbf{0.936} & \textbf{0.936}\\
\multicolumn{1}{c|}{} & Binary-F1  & 0.937 & 0.944 & 0.956 & 0.936 & 0.950 & 0.951 & 0.948 & \underline{0.962} & \textbf{0.963} & \textbf{0.963}\\  \hline

\multicolumn{1}{c|}{\multirow{4}{*}{Slashdot}} & AUC & 0.740 & 0.849 & 0.886 & 0.813 & 0.870 & 0.865 & 0.783 & \underline{0.900} & 0.891 & \textbf{0.902}\\   
\multicolumn{1}{c|}{} & Macro-F1  & 0.688 & 0.729 & 0.780 & 0.667 & 0.750 & 0.745 & 0.683 & \textbf{0.792} & 0.785 & \underline{0.789} \\
\multicolumn{1}{c|}{} & Micro-F1  & 0.786 & 0.823 & 0.855 & 0.813 & 0.842 & 0.833  & 0.811 & \textbf{0.864} & 0.859 & \underline{0.863}\\
\multicolumn{1}{c|}{} & Binary-F1  & 0.869 & 0.889 & 0.908 & 0.887 & 0.902 & 0.895 & 0.884 & \textbf{0.915} & 0.911 & \textbf{0.914}\\  \hline

\Xhline{2\arrayrulewidth}
\end{tabular}%
}
\caption{Link sign prediction performance. \textbf{Bold} and \underline{underline} indicate the best and the second performance respectively. The performances are the average score of 10 experiments with different seed sets.}
\label{tab:my-table}
\end{table*}

Table 1 shows the statistics of the datasets. BitCoin datasets have relatively small size of nodes and edges. The positive and negative ratios are highly imbalanced. Three of the datasets have nearly 90\% of positives among the edges. Otherwise, Slashdot has a 23\% share of negative edges. The four datasets have various graph sizes and positive ratios. It is suitable for unbiased evaluation.

\subsection{Baselines}
We implemented seven baselines to compare the model performance. There are three signed graph convolutions, three constative learnings, and one signed graph contrastive model. 
\begin{itemize}
\item \textbf{SGCN}\cite{derr2018signed} defines (un)balanced path based on the balanced theory for neighbor aggregation. It is the first signed convolution model but does not consider the edge directions. 
\item \textbf{SDGNN}\cite{huang2021sdgnn} proposed four weight matrices to aggregate the neighbor features according to the edge types adaptively. 
\item \textbf{SDGCN}\cite{ko2022graph} is the first spectral convolution for signed-directed graphs. They proposed a signed magnetic Laplacian to overcome the disadvantage of traditional graph Laplacian.
\item \textbf{GraphCL}\cite{you2020graph} is a graph contrastive model with node and edge augmentations. They randomly perturbed graph structures by dropping or adding edges and nodes. 
\item \textbf{GCA}\cite{zhu2021graph} proposed score-based graph augmentation methods and introduced novel node-level contrastive objectives.
\item \textbf{SimGRACE}\cite{xia2022simgrace} tried to overcome the cumbersome augmentation search. They introduced a novel graph encoder perturbation rather than graph augmentation.
\item \textbf{SGCL}\cite{shu2021sgcl} is a graph contrastive model for signed-directed graphs. It perturbs the edge sign and directions to get positive and negative graph views. Each view makes node representations and defines a node-level contrastive loss.
\end{itemize}
For a fair comparison, all the baselines were implemented under the same environments, such as embedding dimension, learning epochs, and convolution layers. Though the graph contrastive baselines are designed for self-supervised learning, we train them with the same supervised loss. Moreover, we removed the read-out process of GraphCL and SimGRACE, which are designed for graph embedding.

\begin{figure*}[t]
  \centering
  \includegraphics[width=\linewidth]{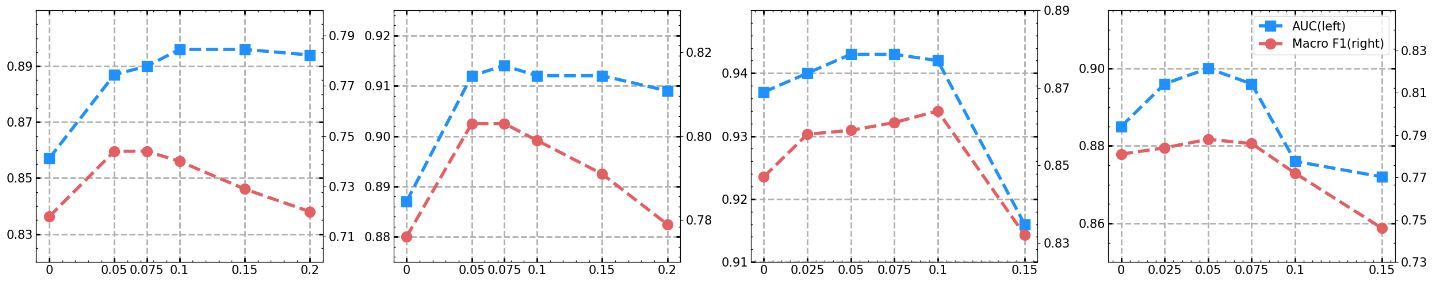}
  \caption{Structure perturbing analysis. x-axis indicates perturbing ratio.}
\end{figure*}

\begin{figure*}[t]
  \centering
  \includegraphics[width=\linewidth]{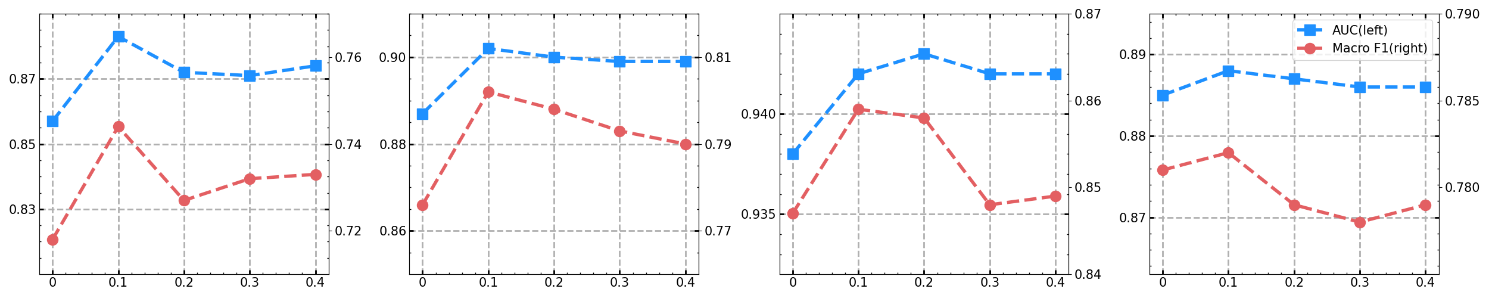}
  \caption{Laplacian perturbing analysis. x-axis indicates noise variance.}
\end{figure*}

\subsection{Implementation Details}
We follow the hyperparameter settings of the original papers of each model. The node embedding dimension is set to 64 for all the baselines to make the same learning capacity. The edges are split into 60:20:20 for training, validation, and test sets. However, we did not use all the positive edges as training instances during the training stage. We sampled positive edges at the ratio of 3:1. It is because the ratio of the positive edge is too high. If we use all positive edges for training, the model would be easy to make positives only. Nevertheless, we utilize all edge instances for the validation and test phase. The structure perturbing ratio $p$ and $r$ are set to 0.1 for all datasets. And the magnetic Laplacian phase $q$ is randomly selected from [0, 0.1$\pi$, 0.2$\pi$, 0.3$\pi$, 0.4$\pi$] for every iterations. The influence of the perturbing ratio and $q$ variation is analyzed in Figure 5. The contrastive loss weight $\alpha$ is set to 0.2. Graph encoder stacks two signed-directed spectral convolution layers. We use Adam optimizer with learning rate = 0.001, weight decay = 0.001. All experiments are run 10 times with different seed sets to avoid randomness and get the average score. The experiments are conducted on Xeon E5-2660 v4 and accelerated via Nvidia Titan XP 12G GPU. The software is implemented via Ubuntu v16.4 with python v3.6 and Pytorch v1.8.0.

\section{Results}
\subsection{Link Sign Prediction}
We summarize the prediction results in Table 2. There are three proposed models. SDGCL-s is a model only with structure augmentation, and SDGCL-l is a model with Laplacian augmentation. SDGCL is the one that adopts both augmentations. Consequently, the proposed SDGCL and its variants always show the best performance in all datasets and all metrics. For Bitcoin-Alpha, Bitcoin-OTC, and Slashdot datasets, SDGCL-composite and SDGCL-structure have the highest score alternatively. SDGCL-composite and SDGCL-Laplacian get the highest score in the Epinions dataset. \\
SGCL or SDGCN shows the best performance among the baselines. SDGCN could achieve better performance than others thanks to the signed-directed spectral convolution. They could fully enjoy the sign and direction information. SGCL is the only model for signed graph contrastive. They could improve the performance via a contrastive learning mechanism than other baselines. However, SGCL shows poor performance on Epinions and Slashdot. We think there are two reasons. First, the advantage of contrastive learning is maximized with scarce labeled datasets. That is why the model can learn more from the contrastive loss. However, if the label is rich enough, the utility of contrastive learning is lessened. Second, we ruin the original dataset context by random perturbation of SGCL. Epinions and Slashdot have larger sizes of edges. The hidden patterns and joint distributions of edges are far more complicated. We harm the original data information by perturbing many edges and splitting them into different graph views. Unlike SGCL, our proposed model minimizes the structure perturbation by proposing Laplacian augmentation. Furthermore, we do not split positive and negative edges in different graph views. 

\subsection{Perturbing Analysis}
Here we analyze the effect of structure and Laplacian augmentation. Figure 4 shows that the performance varies according to the edge perturbing ratio. Laplacian perturbation is not adopted in this experiment to get the effect of edge perturbing. When the perturbing ratios are zero, AUC and Macro-F1 scores are low. We can see the importance of structure perturbing. The performances go up with a small perturbing ratio and go down with a large value. Figure 5 shows the performances of magnetic Laplacian phase $q$ variation. We set default $q$ as 0.1$\pi$ and add some Gaussian noise with standard deviation. At the same time, the edge perturbing ratios are set to zeros. The x-axis shows the standard deviation values. We cap the maximum value of $q$ to $0.5\pi$. Similar to structure perturbation, zero standard deviations show low performances. Zero standard deviation means there is no Laplacian perturbation. They have similar trends that go up and go down. However, Laplacian perturbation is not as sensitive as structure perturbation. 
Though graph augmentations are compulsory in contrastive learning, much perturbation is detrimental to training. Structure perturbation gives direct contrastive information while it may lose the data context. Laplacian perturbation gives indirect perturbation of graph data but is still effective. SDGCL combines these two augmentations for better and more stable graph augmentation.

\subsection{Ablation Study}
Table 3 shows the AUC scores of ablation studies. We check the effects of SDGCL components. w/o Laplacian, w/o structure, and w/o augmentation are the variation of graph augmentations. Especially, w/o augmentation shows the lowest performance. As we expected, the model leverages the advantage of contrastive learning. It shows that augmentations are important in our model. w/o contrastive loss is a model with $\alpha=0$ and the model uses label loss only. Note that it does not mean that the model does not utilize the benefits of contrastive learning. Even though the contrastive loss weight is zero, label loss is calculated with augmented representations of graph views. Thus, the model is still leveraging the contrastive effects. w/o projection is a model without a projection layer. It is well known that the projection layer is useful for robust contrastive learning\cite{jacovi2021contrastive,chen2020big}.

\begin{table}[h]
\centering
\renewcommand{\arraystretch}{1.3}
\resizebox{\columnwidth}{!}{%
\begin{tabular}{lcccc}
\Xhline{2.5\arrayrulewidth}
 & Bitcoin-Alpha & Bitcoin-OTC & Epinions & Slashdot \\ \hline
SDGCL &0.886 & 0.910 & 0.942 & \textbf{0.902} \\
\quad w/o structure aug  & 0.883 & 0.902 & \textbf{0.943} & 0.891 \\
\quad w/o Laplacian aug & \textbf{0.896} & \textbf{0.914} & 0.941 & 0.900 \\
\quad w/o augmentation & 0.846 & 0.889 & 0.936 & 0.884 \\
\quad w/o contrastive loss& 0.878 & 0.901 & 0.941 & 0.891 \\
\quad w/o projection & 0.872 & 0.904 & 0.939 & 0.897 \\
\Xhline{2.5\arrayrulewidth}
\end{tabular}%
}
\caption{Ablation Study}
\end{table}

\section{Conclusion}
This paper proposed SDGCL, a signed-directed graph contrastive learning model. To make augmented node representations, we introduced two levels of the perturbation process. Structure perturbation randomly changes the signs and directions of edges. Even though it may lose some vital information from the original graphs, we check that it makes the model noise-robust. Laplacian perturbation changes the phase parameter $q$ for every training iteration to give perturbation. It does not directly influence the graph information but perturbs the Laplacian matrix. Then we define the inter- and intra-view contrastive objectives with the augmented node representations. 
Unlike other signed graph studies, our model does not depend on social theories or assumptions. Also, there are no arbitrary graph-splitting processes. Our graph encoder is based on the Laplacian matrix; it only enjoys the structural information. Therefore it could reduce computational costs. We evaluate the proposed model with four real-world datasets and seven baselines. The experimental results show that our proposed model has superior performance to others. SDGCL and their variants ranked at the top for every dataset and metric.

\bibliographystyle{ACM-Reference-Format}
\bibliography{reference}

\end{sloppypar}
\end{document}